\documentclass[journal]{IEEEtran}

\usepackage{graphicx}
\usepackage{amsmath,amssymb} 
\usepackage{color}
\usepackage{array}
\usepackage{multirow}

\usepackage{wrapfig,lipsum}
\usepackage{caption}
\usepackage{subcaption}
\usepackage{enumitem}
\usepackage{multirow}
\usepackage{booktabs}
\usepackage{textcomp}
\usepackage{color}
\usepackage{footnote}
\usepackage{empheq}
\usepackage{pifont}
\usepackage[pagebackref=true,breaklinks=true,letterpaper=true,colorlinks,bookmarks=false]{hyperref}

\usepackage{floatrow}

\ifCLASSINFOpdf
\else
\fi

\begin{document}
%
\title{SCAN: Self-and-Collaborative Attention Network for Video Person Re-identification}

\author{Ruimao Zhang, Jingyu Li, Hongbin Sun, Yuying Ge,  Ping Luo, Xiaogang Wang, and Liang Lin\IEEEcompsocitemizethanks {\IEEEcompsocthanksitem
Ruimao Zhang, Ping Luo and Xiaogang Wang are with the Department of Electronic Engineer, The Chinese University of Hong Kong, Hong Kong,  P. R. China (E-mail: ruimao.zhang@cuhk.edu.hk; pluo@ie.cuhk.edu.hk; xgwang@ee.cuhk.edu.hk).
\IEEEcompsocthanksitem
Jingyu Li, Hongbin Sun and Yuying Ge are with Sensetime Research, Shenzhen, P. R. China (E-mail: \{lijingyu,sunhongbin,geyuying\}@sensetime.com).
\IEEEcompsocthanksitem
Liang Lin is with the School of Data and Computer Science, Sun Yat-sen University, Guangzhou, P. R. China (E-mail: linliang@ieee.org).}
}


\markboth{IEEE Transactions on Image Processing}%
{Shell \MakeLowercase{\textit{et al.}}: Bare Demo of IEEEtran.cls for IEEE Journals}

\maketitle

\begin{abstract}

Video person re-identification attracts much attention in recent years.
It aims to match image sequences of pedestrians from different camera views.
Previous approaches usually improve this task from three aspects, including a) selecting more discriminative frames, b) generating more informative temporal representations, and c) developing more effective distance metrics.
To address the above issues, we present a novel and practical deep architecture for video person re-identification termed Self-and-Collaborative Attention Network (SCAN),
which adopts the video pairs as the input and outputs their matching scores.
SCAN has several appealing properties.
First, SCAN adopts non-parametric attention mechanism to refine the intra-sequence and inter-sequence feature representation of videos,
and outputs self-and-collaborative feature representation for each video,
making the discriminative frames aligned between the probe and gallery sequences.
Second, beyond existing models, a generalized pairwise similarity measurement is proposed to generate the similarity feature representation of video pair by calculating the Hadamard product of their self-representation difference and collaborative-representation difference.
Thus the matching result can be predicted by the binary classifier.
Third, a dense clip segmentation strategy is also introduced to generate rich probe-gallery pairs to optimize the model.
In the test phase, the final matching score of two videos is determined by averaging the scores of top-ranked clip-pairs.
Extensive experiments demonstrate the effectiveness of SCAN, which outperforms top-1 accuracies of the best-performing baselines on iLIDS-VID, PRID2011 and MARS dataset, respectively.

\end{abstract}


\IEEEpeerreviewmaketitle

\section{Introduction}


\begin{figure}[t]
\centering
\includegraphics[width=\linewidth]{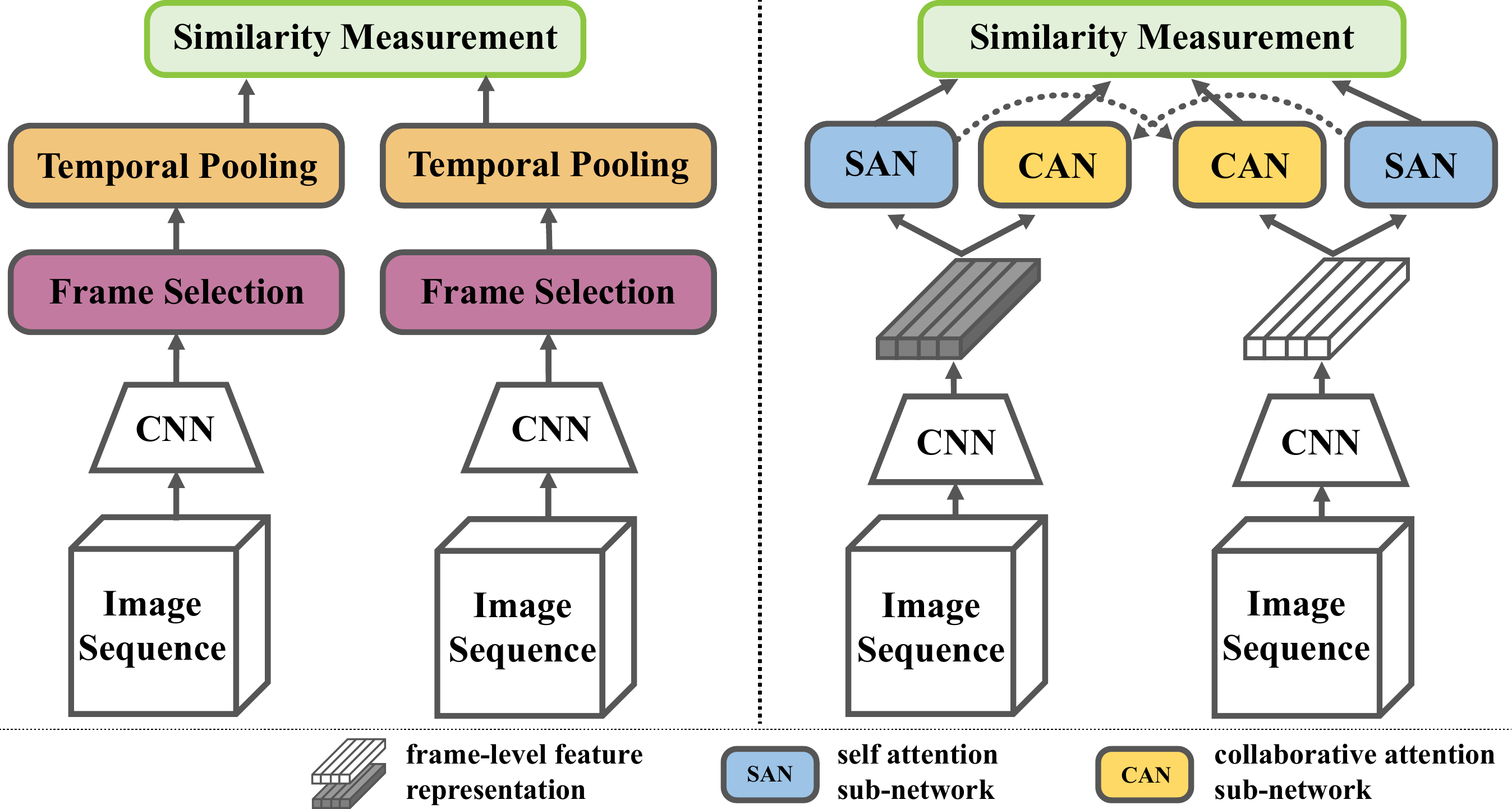}
\caption{ A comparison between a standard deep video re-identification pipeline (left) and the proposed Self-and-Collaborative Attention Network (right).
For example, in the left, the CNN adopts image sequence of two person as input and outputs the feature representation of each frames for both probe and gallery identity.
Then frame selection and temporal pooling are carried out in turn.
Similarity measurement of two identities are conducted at last.
In contrast, our method contains two kinds of frames selection modules. The self attention subnetwork (SAN) is used to select frames from the sequence itself to enhance its feature representation,
and the collaborative attention subnetwork (CAN) is used to select frames from probe or gallery sequence based on the representation of the other one.
The sequence-level representation is also generated in SAN and CAN.
The similarity feature of two input identities are computed in a similarity measurement module according to the outputs of SAN and CAN.}
\label{fig:property}
\end{figure}


As one of the core problems in intelligent surveillance and multimedia application, person re-identification attracts much attention in literature~\cite{C:RelativeDis,C:LADF,C:LFDA,C:CNN-RNN,A:DeepRNN,C:JSTR,J:View-Specific}.
It aims to re-identify individual persons across non-overlapping cameras distributed at different physical locations.
In practice, dramatic appearance changes caused by illumination, occlusions, viewpoint and background clutter increases the difficulty of re-id task.
A lot of work have been proposed to deal with these problems in still images~\cite{C:RelativeDis,C:LADF,C:LFDA,C:DiscriNull,C:SpatialConst,C:LatentPart,J:Bit-scalable}.
Beyond this, there also exist several studies~\cite{C:CNN-RNN,A:DeepRNN,C:JSTR,C:ASTPN,C:DiversReg} discussing the re-id task under image sequence (video) setting.
Since an image sequence usually contains rich temporal information, it is more suitable to identify a person under complex environment and large geometric variations.

As shown in Fig.\ref{fig:property}, besides extracting the feature representation of each frame by convolutional neural networks (CNN),
existing deep video re-identification methods usually include following steps:
a) selecting the discriminative video frames from probe and gallery video sequences respectively,
b) generating informative temporal representation of each video,
c) using video representations and learned similarity measurement to rank the video sequences in the gallery set.
Most previous studies only pay attention to one or two above steps independently.

On the other hand, inspired by~\cite{C:AttentionAll}, several studies~\cite{C:JSTR,C:ASTPN} introduced the attention mechanism to video re-id task for frame selection and temporal modeling.
For example,  Xu \textit{et al}.~\cite{C:ASTPN} adopted the attention matrix to jointly extract the discriminative frames from probe and gallery videos,
and calculated the coupled probe and gallery video representations by temporal pooling operation.
Although these methods achieved promising results, the attention networks are still not fully and effectively explored for temporal modeling.
It can be summarized in the following aspects.
First, existing methods usually generate the sequence representations after aligning probe and gallery frames (\textit{i.e.} assigning attention weight to each frame),
thus the video representation cannot in turn refine the frame selection.
For example, if the probe video from person-1 has some occlusion frames with the similar appearance to the gallery video from person-2,
these occlusion frames will obtain great attention in the frame alignment, and further affect final video representations.
Second, some combinations of similarity measurements and loss functions in previous studies are not suitable for the attention mechanism to discover discriminative frames.
Third, the attention mechanism in existing methods is usually parametric, making the length of the input sequence or the feature dimension of frames need to be fixed.

In order to address the above issues,
we propose a simple but effective architecture termed Self-and-Collaborative Attention Network (SCAN) to jointly deal with frame selection, temporal modeling and similarity measurement for video re-identification task.
As shown in Table~\ref{table:contribution}, it has several benefits that existing methods do not have.
a) Compared with the recurrent neural network (RNN) based attention models, SCAN adopts attention mechanism to refine the intra-sequence and inter-sequence feature representation of videos.
Such process can efficiently align discriminative frames between the probe and gallery image sequences.
The output self and collaborative feature representations leverage the global temporal information and local discriminative information.
b) We propose a generalized pairwise similarity measurement in SCAN, which adopts self and collaborative video representations to calculate the similarity features of video-pairs.
Thus the matching problem can be transformed into a binary classification problem, and the label of an identity pair is used to optimize the classifier.
Such module encourages the video features from the same identity to be similar, and enlarges the distance between informative frames and noisy frames in the same video.
Moreover, different from pair-wise loss or triplet loss that needs a predefined margin constraint~\cite{C:quadrupletloss}, the binary loss can reduce the cost to tune such hyperparameter.
c) The attention module in SCAN is non-parametric, thus it can deal with image sequence with various lengths and the input feature dimensions of each frame are also variable .
d) A dense clip segmentation strategy is introduced to generate much more probe-gallery pairs (including the hard positive and hard negative pairs) to optimize the model.

As shown on the right of Fig.\ref{fig:property}, in practice, we first extract the feature representation of each frame (black and white rectangles) from both probe and gallery videos using pre-trained CNN.
Then we input the frame-level feature representations from the probe and gallery videos into self attention subnetwork (SAN) independently,
After calculating the correlation (the attention weight) between the sequence and its frames,
the output sequence representation is reconstructed as a weighted sum of the frames at different temporal positions in the input sequence.
%
%
We also introduce the collaborative attention subnetwork (CAN) to calculate the coupled feature representations of the input sequence pair.
The calculation process of CAN is the same as the SAN, but the meaning of the output varies according to different inputs.
For instance, if the input sequence-level feature is from the probe video and the frame-level features are from the gallery video, the output of CAN will be the probe-driven gallery representation. Otherwise, it will be the gallery-driven probe representation.
After SAN and CAN, we calculate the difference between self-representations of probe and gallery videos, as well as the difference between their collaborative-representations.
These two differences are merged by the Hadamard product and fed into a fully-connected layer to calculate the final matching score.
%


\begin{table}[t]
\begin{center}
\caption{  The proposed SCAN integrates the benefits of the previous work into a unique framework, and also introduces some elaborate mechanisms to further improve the performance of video re-id.
`Frame Align.', `P-G Inter.', `Var. Dim.' and `P-G Pair Aug.' are short for frame alignment, probe-and-gallery interaction, accepting temporal modeling with various feature dimensions (i.e. various number of frames and feature channels) and probe-and-gallery pair augmentation. }
\label{table:contribution}
\begin{tabular}{ l || p{0.9cm}<{\centering} | p{0.9cm}<{\centering} | p{0.9cm}<{\centering} | p{0.9cm}<{\centering} }
\hline
 Method  & Frame Align.    & P-G Inter.  &   Var. Dim. & P-G  Aug. \\
\hline
Mc. \textit{et al}.~\cite{C:CNN-RNN}             &             &            &  \checkmark &    \\
Zhou \textit{et al}.~\cite{C:JSTR}               &             & \checkmark &              &    \\
Xu \textit{et al}.~\cite{C:ASTPN}                & \checkmark  & \checkmark &              &   \\
Liu \textit{et al}.~\cite{C:QAN}                 &             &            &  \checkmark &       \\
Li \textit{et al}.~\cite{C:DiversReg}            &             &            &   &  \\
\hline
our method                                       & \checkmark &  \checkmark &  \checkmark & \checkmark   \\
\hline
\end{tabular}
\end{center}
\end{table}

\begin{table*}[t]
\begin{center}
\caption{Comparisons between proposed SCAN and other state-of-the-arts for video person re-id.
$\checkmark$ represents the methods or information indicated by the column indices are adopted.
`non-para.' is short for non-parametrization in temporal modeling. `P-G. Inter.' denotes probe-gallery interactions during sequence representation generation.
The abbreviations `attention', `pooling' in the third column represent attention mechanism and pooling operation.
The uppercase `P',`T' and `B' in the sixth column indicate pairwise loss, triplet loss and binary loss, respectively.
In the last column, `cons.' denotes the clip of each video is extracted from consecutive frames,
and `rand.' means randomly extracting several frames from the video as the clip.
The `dense' indicates our model segment the image sequence into multiple clips for model training.
}
\label{table:method}
\begin{tabular}{ l || p{1cm}<{\centering} | p{4.3cm}<{\centering} | p{1.4cm}<{\centering} | p{1cm}<{\centering} | p{1cm}<{\centering} |  p{1.2cm}<{\centering} | p{0.9cm}<{\centering}  }
\hline
\multirow{2}{*}{Method } &   Spatial & \multicolumn{2}{c|}{ Temporal Modeling  } & P-G.  & Loss  & Identity & Video \\
\cline{3-4}
                         &   Info.    &  method & non-para.                    & Inter. & Func.& Loss & Clips\\
\hline
Mc. \textit{et al}.~\cite{C:CNN-RNN}  & & RNN~+~pooling & &  &P &  \checkmark & cons.     \\
Zhou \textit{et al}.~\cite{C:JSTR}     & \checkmark & LSTM~+~attention~+~pooling& & \checkmark & T,~B &     &   rand.  \\
Xu \textit{et al}.~\cite{C:ASTPN}      & \checkmark & RNN~+~attention~+~pooling& & \checkmark & P &  \checkmark  &  cons.    \\
Liu \textit{et al}.~\cite{C:QAN}       & & weighted sum & & & T &  \checkmark &      \\
Zhong \textit{et al}.~\cite{C:k-reciprocal}      &  & max pooling & \checkmark &\checkmark & P& \checkmark   &    \\
Li \textit{et al}.~\cite{C:DiversReg}          &\checkmark    &  weighted sum     &    &   & --  & \checkmark  &   rand.                 \\
\hline
our method                & & SCAN & \checkmark &\checkmark & B &  \checkmark &  dense   \\
\hline
\end{tabular}
\end{center}
\end{table*}

In general, the \textbf{contribution} of this work can be summarized in three folds.
\begin{itemize}

\item We propose a Self-and-Collaborative Attention Network (SCAN) to efficiently align the discriminative frames from two videos.
It includes a non-parametric attention module to generate self and collaborative sequence representations by refining intra-sequence and inter-sequence features of input videos,
and a generalized similarity measurement module to calculate the similarity feature representations of video-pairs.

\item We introduce such a module into video re-identification task, and propose a novel and practical framework to simultaneously deal with frame selection, video temporal representation and similarity measurement.
In addition, a dense clip segmentation strategy is also introduced to generate much more probe-gallery pairs to optimize the model.

\item  The proposed model outperforms the state-of-the-art methods on top-1 accuracy in three standard video re-identification benchmarks.

\end{itemize}

The rest of the paper is organized as follows. Section~\ref{sec:Relatedwork} presents a brief review of related work. Section~\ref{sec:methodlogy} introduces our Self and Collaborative Attention Network. The experimental results, comparison and component analysis are presented in Section~\ref{sec:experiment}. Section~\ref{sec:conclusion} concludes the paper.

\begin{figure*}[t]
\centering
\includegraphics[width=\linewidth]{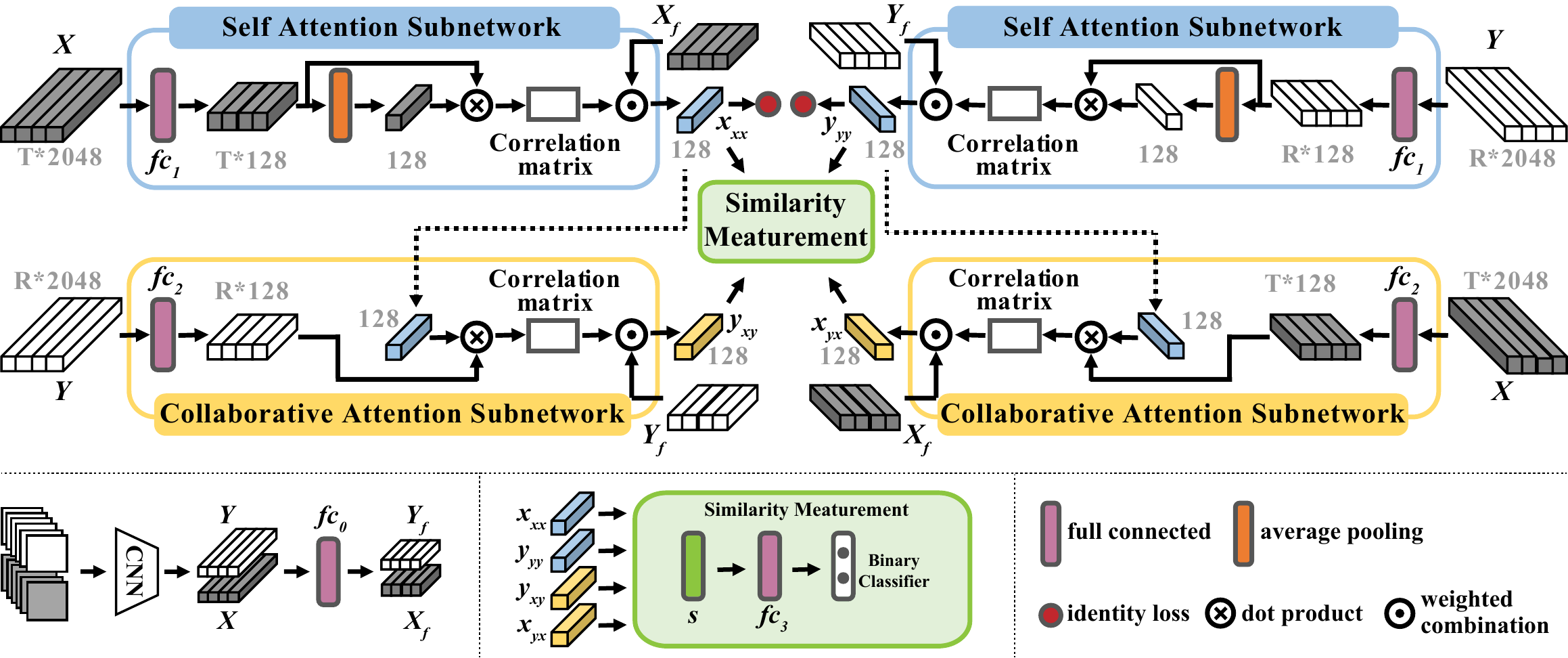}
\caption{Architecture of proposed Self-and-Collaborative Attention Network for video-based person re-identification.
This architecture is comprised of four parts: shared convolution neural networks, self attention subnetwork (SAN), collaborative attention subnetwork (CAN) and similarity measurement module.
For each person in probe and gallery set, the video clips extracted from their image sequences are first fed into CNN to obtain frame-level features.
Then SAN and CAN are adopted to generate sequence-level representations according to the non-parametric attention mechanism.
At last, the binary cross-entropy loss and identity loss are used to optimize the parameters of SCAN.
Zoom in four times for best view.}
\label{fig:framework}
\end{figure*}

\section{Related Work}
\label{sec:Relatedwork}


\noindent
\textbf{Person re-identification}.
Person re-id in still image has been extensively explored in the literature~\cite{C:RelativeDis,C:LADF,C:LFDA,C:MaxOccu,C:DiscriNull,C:SpatialConst,C:LatentPart,C:k-reciprocal,C:Group-Shuffling,C:KroneckerProduct}
in the past few years.
Traditional methods were mainly concentrated on the hand-craft appearance feature design according to the human domain knowledge~\cite{J:ColorInvariants,C:symmetry-driven,C:ViewpointLocal,C:HierGaussian,C:CustomPictorial},
since ideal feature representation can be sufficiently against camera view changes~\cite{J:CameraCorrelation}.
Another crucial component for person re-identification is the distance metric learning~\cite{J:Metric-jmlr,C:InforMetric,C:deepMetric,C:Transitive,J:GenerSimil,C:Normface}, which is applied to learn a common space for data from different domains. .
%
%
The other approaches paid much attention on view-specific learning mechanism which learned the individual matching weights for each camera view. Some CCA-based methods~\cite{C:ROCCA,J:CVDCA} belonged to this category.

With the emergence of deep learning, several distance-driven feature learning frameworks have been proposed via adopting Convolutional Neural Networks (CNN) to jointly learn similarity measurement and discriminative feature representations for person re-identification~\cite{J:DeepRelative,C:multi-channel,C:quadrupletloss,C:GuidedDrop,C:Multi-scaleRep,A:yuncong,C:deepreid}.
Combined with these frameworks, attention mechanism has also been widely applied in the re-identification problem.
Notable models included attention to multi-level features~\cite{C:FactNet,C:Multi-Level2018}, attention to discriminative pixels or regions~\cite{J:ComparativeAttention,C:KroneckerProduct,C:HarmoniousAttention}.
Besides the above methods, there also exist some studies incorporating the affinities between gallery images into the re-ranking process to further improve the matching accuracy~\cite{C:post-rank,C:RankingOptim,C:RankingAggregation,J:post-rankingTIP,C:k-reciprocal,C:Group-Shuffling}.
For example, in~\cite{C:k-reciprocal}, Zhong \textit{et al.} used the $k$-reciprocal neighbors of probe and gallery images to calculate the Jaccard distance to further re-rank the matching list.
Motivated by this work, Shen \textit{et al.}~\cite{C:Group-Shuffling} further integrated the above process into model training, and adopted gallery-to-gallery affinities to refine the probe-to-gallery affinity with a simple matrix operation.

Recently, the studies about video-based person re-identification adopted image sequence to further improve the matching accuracy~\cite{C:CNN-RNN,C:MARS,A:DeepRNN,C:JSTR,C:TwoStreamReId,J:I2V,C:ASTPN}.
For example, McLaughlin \textit{et al.}~\cite{C:CNN-RNN} proposed a basic pipeline for deep video re-id. It uses CNN to extract the feature of each frame.
Then the RNN layer is applied to incorporate temporal context information into each frame,
and the temporal pooling operation is adopted to obtain the final sequence representation.
Both the identity loss and siamese loss are used to optimize parameters.
In~\cite{A:DeepRNN}, Wu \textit{et al}. proposed a similar architecture to jointly optimize CNN and RNN to extract the spatial-temporal feature representation for similarity measurement.
Same as the still image, one of the remarkable property in recent video-based studies is applying the attention mechanism to discover the discriminative frames from probe and gallery videos.
As shown in Table~\ref{table:method},
Zhou \textit{et al}.~\cite{C:JSTR} proposed a temporal attention mechanism to pick out the discriminative frames for video representation.
Moreover, the spatial RNN is adopted to integrate the context information from six directions to enhance the representation of each location in the feature maps.
Li \textit{et al}.~\cite{C:DiversReg} proposed a spatiotemporal attention model and diversity regularization to discover a set of distinctive body parts for the final video representation.
In~\cite{C:ASTPN}, Xu \textit{et al}. introduced the shared attention matrix for temporal modeling, realizing the information exchange between probe and gallery sequence in the process of frame selection.
In such case, the discriminative frames can be aligned according to their attention weights.

The proposed SCAN is partially related to the above methods~\cite{C:CNN-RNN,A:DeepRNN,C:JSTR,C:DiversReg,C:ASTPN}, which adopt the attention mechanism to extract the rich spatial-temporal information for feature representation.
But the technical details of these work are different from our method.
Since the proposed SCAN outputs the attention weights by leveraging  global temporal information and local discriminative information,
it is more robust to deal with the noise frames during alignment.
On the other hand, it is a non-parametric module, thus can be more flexible to deal with various sequence lengthes and feature dimensions.
In~\cite{C:DualAttention}, Si \textit{et al.} also applied the non-parametric attention module for feature refinement and feature-pair alignment.
The different between this method and our SCAN are three folds. First, their method applied frame-level feature refinement and the temporal modeling was omitted.
Second, they adopted Euclidean distance to measure the similarity between the inter-sequence refined features and intra-sequence refined features, but we proposed a novel similarity feature to represent the difference between probe and gallery sequences.
At last, we replace the triplet loss in their method with cross-entropy loss to optimize the model.

%

\noindent
\textbf{Self-attention and interaction network}.
Recent developed self-attention~\cite{C:AttentionAll} mechanism for machine translation is also related to our work.
It calculated the response at one position as a weighted sum of all positions in the sentence.
The similarity idea was also introduced in Interaction Networks (IN)~\cite{C:IN1,C:IN2,C:VAIN} for modeling the pairwise interactions in physical systems.
Recently, Wang \textit{et al}.~\cite{C:NonLocal} extended these methods into computer vision area, and proposed the Non-Local Network to model the long-range spatial and temporal dependencies in a single block.
In~\cite{A:TRN}, Zhou \textit{et al}. proposed the Temporal Relation Network (TRN) to learn temporal dependencies between video frames at multiple time scales.
The proposed SCAN is inspired by above two works, but we further introduce the collaborative representation mechanism to deal with the matching problem.

\noindent
\textbf{Collaborative representation}. Learning collaborative representation aims to represent a sample as a weighted linear combination of all training samples. It has been successfully applied in many computer vision tasks, such as face recognition~\cite{J:SparseCoding,C:SRCR}, super-resolution~\cite{J:SuperResolution}, and image denoising~\cite{C:denoising}.
In this paper, we introduce a collaborative representation into temporal modeling, and combine it with deep neural networks for end-to-end training.
Specially, self and collaborative attention network are proposed to represent the video as a weighted combination of multiple frames. It is a non-parametric method that can effectively align the discriminative frames in probe and gallery videos.

\section{Methodology}
\label{sec:methodlogy}

%
%

\subsection{Deep Architecture}
\label{sec:architecture}

Given a query video $\textbf{I} = \{I^t \}_{t=1}^T$ (\textit{i.e.} an image sequence) of a person, where $T$ is the number of frames in the video,
the target of video-based person re-identification is to find the correct matching result among a large number of candidate videos extracted from the different camera views.
In other words, for the video $\textbf{I}_A$ of person $A$, we wish to learn a sequence-based re-identification model to distinguish whether or not another video $\textbf{I}_{A^*}$ captures the same person from other space or time.
In the following, we firstly give the framework of proposed SCAN, and then describe technique details of its modules.

\textbf{Feature extraction}.
Same as previous work~\cite{C:RelativeDis,C:CNN-RNN,C:MARS}, all of the image sequences are divided into a gallery set and a probe set.
The gallery set is usually consisted of one image sequence for each person captured from a special camera view, and the other image sequences are employed as the probe set.
The performance of a video re-identification model is evaluated according to the locations of the correctly matched probe videos in the ranking list of corresponding gallery videos.
The deep architecture of proposed method is illustrated in Fig.~\ref{fig:framework}.
Supposing the probe image sequence is represented as $\textbf{I}_p = \{I_p^t \}_{t=1}^T$ and  the gallery sequence is as $\textbf{I}_g = \{I_g^r \}_{r=1}^R$. $T$ and $R$ indicate the length of the image sequences.
The probe and gallery sequences are at first fed into CNN to extract the feature representation of each frame. The parameters of CNN are shared for both sequences.
Let the feature representation of the probe and gallery person be $\textbf{X} = \{\textbf{x}^t | \textbf{x}^t \in \mathbb{R}^{d}  \}_{t=1}^T$ and $\textbf{Y} = \{\textbf{y}^r | \textbf{y}^r \in \mathbb{R}^{d}\}_{r=1}^R$, where $d$ is the dimension of the feature vector and is set as $2048$ in practice.
We further apply the \textit{fc-0} layer to reduce the feature dimension to $128$ and denote them as $\textbf{X}_f = \{\textbf{x}_f^t  \}_{t=1}^T$ and $\textbf{Y}_f = \{\textbf{y}_f^r \}_{r=1}^R$, respectively.

\textbf{Self Attention Subnetwork}. After feature extraction, the Self Attention Network (SAN) is adopted to select the informative frames  to further enhance the representation of image sequence for each person.
We first feed $\{ \textbf{X}, \textbf{X}_f \}$ and $\{ \textbf{Y},\textbf{Y}_f \}$ into SAN.
Then the dimension of $\textbf{X}$ and $\textbf{Y}$ is reduced from $2048$ to $128$ using  \textit{fc-1} layer and denoted as $\textbf{X}_s = \{\textbf{x}_s^t \}_{t=1}^T$ and $\textbf{Y}_s = \{\textbf{y}_s^r \}_{r=1}^R$.
After that, the sequence-level representation of $\textbf{X}_s$ and $\textbf{Y}_s$ are produced through \textit{average pooling} over the temporal dimension.
Let $\hat{\textbf{x}}_s$ and $\hat{\textbf{y}}_s$ be the sequence-level feature vector of probe and gallery video in SAN, we further enhance these feature representations by,
\begin{equation}\label{eq:scr}
\hat{\textbf{x}}_{xx} = \sum_{t=1}^T f( \textbf{x}_s^t, \hat{\textbf{x}}_s) \circ \textbf{x}_f^t ~~~~~
\hat{\textbf{y}}_{yy} = \sum_{r=1}^R f( \textbf{y}_s^r, \hat{\textbf{y}}_s) \circ \textbf{y}_f^r
\end{equation}
where $f(.,.)$ is a parameter-free correlation function, which outputs the normalized correlation weight (\textit{i.e.} attention weight) of input features.
It may have various forms~\cite{C:NonLocal}.
In this paper, $f(.,.)$ includes two operations that are Hadamard product and the \textit{softmax} operation along the temporal dimension ($t$ and $r$).
The former is used to calculate the correlation weights, and the latter is adopted to normalize the weight vectors in each dimension.
Such operation is inspired by the recent proposed self-attention module~\cite{C:attnetionisall} and non-local operation~\cite{C:NonLocal} in deep neural network.
Different from non-local operation~\cite{C:NonLocal} which is used to aggregate the representation of each site on the feature maps to refine the feature of a certain location,
the output of SAN is the refined feature representation of the entire video clip,
thus the features in different dimensions may capture the different spatial information for a certain identity.
Through replacing the dot product in~\cite{C:NonLocal} by the Hadamard product to calculate the correlation weights,
our method can reduce the impact of dramatic spatial changes on the calculation of the correlation weight.
$\circ$ indicates the element-wise product.
The subscript $_{xx}$ indicates the probe-driven probe representation, while $_{yy}$ indicates the gallery-driven gallery representation.
The output $\hat{\textbf{x}}_{xx}$ and $\hat{\textbf{y}}_{yy}$ are then passed into collaborative attention subnetwork.

\textbf{Collaborative Attention Subnetwork}. The input of CAN is from two branches. One is the sequence-level representation $\hat{\textbf{x}}_{xx}$ and $\hat{\textbf{y}}_{yy}$ from SAN, and the other is the frame-level representations $\{ \textbf{X}, \textbf{X}_f \}$ and $\{ \textbf{Y},\textbf{Y}_f \}$ from CNN.
Same as SAN, we reduce the dimension of $\textbf{X}$ and $\textbf{Y}$ from $2048$ to $128$ using \textit{fc-2} layer in CAN.
The outputs are $\textbf{X}_c = \{\textbf{x}_c^t \}_{t=1}^T$ and $\textbf{Y}_c = \{\textbf{y}_c^r \}_{r=1}^R$.
Then the cross-camera feature representation can be computed as,
\begin{equation}\label{eq:ccr}
\hat{\textbf{x}}_{yx} = \sum_{t=1}^T f( \textbf{x}_c^t, \hat{\textbf{y}}_{yy}) \circ \textbf{x}_f^t ~~~~~
\hat{\textbf{y}}_{xy} = \sum_{r=1}^R f( \textbf{y}_c^r, \hat{\textbf{x}}_{xx}) \circ \textbf{y}_f^r
\end{equation}
The subscript $_{xy}$ indicates the probe driven gallery representation, and $_{yx}$ is the gallery driven probe representation. The operation in Eqn.(\ref{eq:ccr}) enables probe and gallery video to effectively select frames and corresponding discriminative features from each other.


\textbf{Similarity measurement}. We use the output of SAN and CAN to calculate the similarity feature representation of probe sequence and gallery sequence as follows,
\begin{equation}\label{eq:similarity}
\begin{split}
&\textbf{s} = (\hat{\textbf{x}}_{xx} - \hat{\textbf{y}}_{yy}) \circ (\hat{\textbf{x}}_{yx} - \hat{\textbf{y}}_{xy})  \\
&= ( \hat{\textbf{x}}_{xx} \circ  \hat{\textbf{x}}_{yx} - \hat{\textbf{y}}_{yy} \circ \hat{\textbf{x}}_{yx} ) +
( \hat{\textbf{y}}_{yy} \circ \hat{\textbf{y}}_{xy} - \hat{\textbf{x}}_{xx} \circ \hat{\textbf{y}}_{xy}  ) \\
&= ( \textbf{X}_f \cdot \hat{\textbf{c}}_{xx} \circ \textbf{X}_f \cdot \hat{\textbf{c}}_{yx} - \textbf{Y}_f \cdot \hat{\textbf{c}}_{yy} \circ \textbf{X}_f \cdot \hat{\textbf{c}}_{yx} )  \\
& + ( \textbf{Y}_f \cdot \hat{\textbf{c}}_{yy} \circ \textbf{Y}_f \cdot \hat{\textbf{c}}_{xy} - \textbf{X}_f \cdot \hat{\textbf{c}}_{xx} \circ \textbf{Y}_f \cdot \hat{\textbf{c}}_{xy} )
\end{split}
\end{equation}
where  $\hat{\textbf{c}}_{xx}$, $\hat{\textbf{c}}_{yy}$, $\hat{\textbf{c}}_{xy}$, $\hat{\textbf{c}}_{yx}$ denote the combination coefficient matrices calculated by the non-parameter correlation function $f(.,.)$.
The meaning of subscripts are consistent with that in the sequence-level representation.
The operation $\cdot$ indicates weighted combination along each feature dimension (\textit{i.e.} Hadamard product followed by column summation), and $\circ$ denotes the Hadamard product.
According to Eqn.~\ref{eq:similarity}, the self enhanced features can be thought as gating the collaborative enhanced features.
In other words, the self representations of video-pair modulate their collaborative representations to refine the corresponding discriminative frames and features to calculate the final pair-wise similarity.
Note that $\textbf{s}$ is a vector but not a scalar, which indicates the sequence-level similarity after frame-oriented feature selection.

The above feature representation is then transformed by a fully-connected layer, \textit{i.e.} \textit{fc-3} layer, to obtain the final matching score.
At last, we adopt identity-pair annotations and binary cross-entropy loss to optimize the matching scores.
%
%
%
If the probe video and gallery video present the same person identity, the value of the label is $1$, else it will be $0$.
The same operation is also used in textual-visual matching problem~\cite{C:textual-visual}.



\subsection{Compared with Traditional Metric Learning}
\label{sec:connection}

According to~\cite{J:GenerSimil}, the generalized linear similarity of two feature vectors can be written as,
\begin{equation}\label{eq:generalized}
\begin{split}
\tilde{s} &=      \left[ \textbf{x}^T ~~ \textbf{y}^T  \right]
\left[ \begin{matrix} \textbf{A} & -\textbf{C} \\ -\textbf{D} & \textbf{B} \end{matrix} \right]
\left[ \begin{matrix} \textbf{x} \\ \textbf{y} \end{matrix} \right]  \\
&= (\textbf{x}^T \textbf{A} \textbf{x} - \textbf{y}^T \textbf{D} \textbf{x}) + (\textbf{y}^T \textbf{B} \textbf{y} - \textbf{x}^T \textbf{C} \textbf{y}) \\
&= \underbrace{[ (\widetilde{\textbf{A}}\textbf{x})^T~\widetilde{\textbf{A}}\textbf{x} - (\widetilde{\textbf{D}}_y\textbf{y})^T~\widetilde{\textbf{D}}_x\textbf{x} ]  }_{\text{ \textit{Part A} }} \\
&+ \underbrace{ [ (\widetilde{\textbf{B}}\textbf{y})^T~\widetilde{\textbf{B}}\textbf{y} - (\widetilde{\textbf{C}}_x\textbf{x})^T~\widetilde{\textbf{C}}_y\textbf{y} ]  }_{\text{ \textit{Part B} }}
\end{split}
\end{equation}
where $\textbf{A},\textbf{B},\textbf{C}$ and $\textbf{D}$ are the parameters to be optimized, and $\textbf{A}=\widetilde{\textbf{A}}^T\widetilde{\textbf{A}}$, $\textbf{B}=\widetilde{\textbf{B}}^T\widetilde{\textbf{B}}$,
$\textbf{C}=\widetilde{\textbf{C}}_x^T\widetilde{\textbf{C}}_y$ and $\textbf{D}=\widetilde{\textbf{D}}_y^T\widetilde{\textbf{D}}_x$.
When $\textbf{A}=\textbf{B}=\textbf{C}=\textbf{M}$ and $\textbf{D}=\textbf{M}^T$, it degenerates into Mahalanobis distance with the form $\tilde{s} = (\textbf{x}-\textbf{y})^T \textbf{M} (\textbf{x}-\textbf{y})$.

Intuitively, Eqn.(\ref{eq:similarity}) has a very similar form with Eqn.(\ref{eq:generalized}).
The differences are three folds:
First, we replace $\widetilde{\textbf{A}}$, $\widetilde{\textbf{D}}$ and $\widetilde{\textbf{B}}$, $\widetilde{\textbf{C}}$ in \textit{Part A} and \textit{Part B} with two sets of frame-level feature representations $\textbf{X}_f$ and $\textbf{Y}_f$, respectively.
Second, the feature vector $\textbf{x}$ and $\textbf{y}$  in Eqn.(\ref{eq:generalized}) is replaced by the combination coefficients $\hat{\textbf{c}}$ in Eqn.(\ref{eq:similarity}), which are computed by correlation function $f(\cdot,\cdot)$.
For simplicity, we omit the subscript of $\hat{\textbf{c}}$.
Each column of $\hat{\textbf{c}}$ is corresponding to the attention weights for each frames.
At last, some dot product operations are replaced by element-wise product.
The output of the Eqn.(\ref{eq:generalized}) is the matching score, while the Eqn.(\ref{eq:similarity}) generates the similarity feature representation which is further fed into binary classifier for prediction.

%

By making an analogy with the form of proposed similarity measurement and generalized linear similarity,
we provide insight by relating the attention-based similarity module to the previous deep metric learning approach (\textit{i.e.} parameterized similarity module).
Eqn.(\ref{eq:similarity}) and Eqn.(\ref{eq:generalized}) have the similar forms, but their meanings are different.
For the generalized linear similarity in Eqn.(\ref{eq:generalized}), it projects the feature representation $\textbf{x}$ and $\textbf{y}$ into a common feature space by using linear transformations.
$\widetilde{\textbf{A}}$, $\widetilde{\textbf{D}}$ and $\widetilde{\textbf{B}}$, $\widetilde{\textbf{C}}$  are the parameters that need to be optimized.
In contrast to this, our method uses the temporal attention weights $\hat{\textbf{c}}$ to select the  discriminative frames in the probe and gallery videos to generate the final similarity feature of the video pair.
Such operation can be regarded as projecting two image sequences into a common `\textit{frame space}' to align the discriminative frames.
Such a scheme is significant in the temporal-based matching problem, and can also be adopted as a common technique in a series of video-based applications.
%
%
In this sense, our work bridges the generalized deep metric learning with the temporal frame selection strategy.
In addition, it provides a more intuitive perspective to understand the meaning of proposed similarity measurement.

\subsection{Implementation Details}
\label{sec:implementation}

\noindent
\textbf{Clip Segmentation}. In practice, we segment every image sequence into several video clips.
The length of each clip is $10$ and the segmentation stride is set as $5$ in training and test procedure.
When the frames at the end of the video are not sufficient to generate the clip, we discard the rest frames directly.
The advantages of such pre-processing strategy are as follows:
(a) It can generate a large amount of probe-gallery pairs to optimize network parameters, which is critical for the deep model training.
Specially, it is beneficial to produce much more hard positive/negative training pairs to promote the training effiency.
(b) It avoids loading the entire image sequence into the model for temporal modeling.
In such case, when the batch size is fixed, it can increase the diversity of minibatch effectively.
This ensures the training process more stable and BatchNorm (BN)~\cite{C:BN} more efficient to accelerate the model convergence.
In the test phase, we select $10\%$ clip pairs with the highest matching score from coupled image sequences and average their matching scores as the final confidence score.
We rank all of the confidence scores and return the final ranking list to calculate the matching accuracy.
It is worth noting that the re-ranking technique, such as~\cite{C:k-reciprocal}, is omitted in this paper.

\textbf{Training process}.
%
All of the CNN models in this work are pre-trained on ImageNet~\cite{C:ImageNet}.
We fine-tune the models using $16$ identities in each batch. For each identity, we randomly load $2$ video clips for training.
Thus, there are $32$ clips with $320$ video frames as the input for each iteration.
The input frames are resized into $256\times128$ pixels.
Horizontal flipping is also used for data augmentation.
We adopt Online Instance Matching (OIM)~\cite{C:personsearch} loss as the identity loss function.
We train our models on $4$-GPU machine. Each model is optimized $30$ epoches in total, and the initial learning rate is set as $0.001$.
The learning rate is updated with the form, $lr = lr_0 \times 0.001^{(epoch / 10)}$, where $lr_0$ denotes the initial learning rate.
We use a momentum of $0.9$ and a weight decay of $0.0001$.
The parameters in BN layers are also updated in the training phase.


\section{Experiments}
\label{sec:experiment}

%

\begin{table*}[t]
\begin{center}
\caption{Performance comparison on the iLIDS-VID by state-of-the-art methods. Our model is based on ResNet50. Top-1, -5, -10, -20 accuracies(\%) are reported.}
\label{table:iLIDS-VID}
\begin{tabular}{| l| c | p{0.8cm}<{\centering} | p{2.2cm}<{\centering} |p{0.9cm}<{\centering} |p{1.3cm}<{\centering} p{1.3cm}<{\centering} p{1.3cm}<{\centering} p{1.3cm}<{\centering}|  }
\hline
 \multirow{2}{*}{Methods} & \multirow{2}{*}{Reference} & Deep  & Backbone & Optical     &   \multicolumn{4}{c|}{  iLIDS-VID } \\
\cline{6-9}
                          & & model  & Network & Flow    &    top-1 & top-5 & top-10 & top-20    \\
\hline
1.~LFDA~\cite{C:LFDA}     & cvpr13 & no & -- & no & 32.9  & 68.5 & 82.2 & 92.6 \\
2.~LADF~\cite{C:LADF}     & cvpr13 & no & -- & no & 39.0  & 76.8 & 89.0 & 96.8 \\
3.~STFV3D~\cite{C:STFV3D} & iccv15 & no & -- & no & 44.3 & 71.7 & 83.7 & 91.7  \\
4.~TDL~\cite{C:TDL}       & cvpr16 & no & -- & no & 56.3 & 87.6 & 95.6 & 98.3 \\
5.~CNN-RNN~\cite{C:CNN-RNN}&cvpr16 & yes& SiameseNet~\cite{C:SiameseNet} & yes & 58.0 & 84.0 & 91.0 & 96.0 \\
6.~CNN+XQDA~\cite{C:MARS} & eccv16 & yes& CaffeNet&no  & 53.0 & 81.4 & --& 95.1 \\
7.~TAM+SRM~\cite{C:JSTR}  & cvpr17 & yes& CaffeNet&no & 55.2 & 86.5 & -- & 97.0 \\
8.~ASTPN~\cite{C:ASTPN}   & iccv17 & yes& SiameseNet~\cite{C:ASTPN} & yes & 62.0 & 86.0 & 94.0 & 98.0 \\
9.~QAN~\cite{C:QAN}       & cvpr17 & yes& RPN~\cite{C:RPN} &no & 68.0 & 86.8 & 95.4 & 97.4\\
10.~RQEN~\cite{C:RQEN}    & aaai18 &yes & GoogLeNet&no  & 77.1 & 93.2 &\textcolor{blue}{\textbf{97.7}} & \textcolor{blue}{\textbf{99.4}} \\
11.~STAN~\cite{C:DiversReg}&cvpr18 &yes &  ResNet50 &no & 80.2 &  --  & --  & -- \\
12.~ST-Tubes~\cite{A:ST-Tubes}.&arxiv19&yes& ResNet50 &no& 67.0 & 84.0& 91.0&96.0 \\
\hline
12.~ours w/o optical     & -- & yes &  ResNet50 &no  & \textcolor{blue}{\textbf{81.3}} & \textcolor{blue}{\textbf{93.3}} & 96.0 & 98.0\\
13.~ours w/ optical      & -- & yes & ResNet50 & yes  & \textcolor{red}{\textbf{88.0}} & \textcolor{red}{\textbf{96.7}} & \textcolor{red}{\textbf{98.0}} & \textcolor{red}{\textbf{100.0}} \\
\hline
\end{tabular}
\end{center}
\end{table*}

\begin{table*}[t]
\begin{center}
\caption{  Performance comparison on the PRID2011 by state-of-the-art methods. Our model is based on ResNet50. Top-1, -5, -10, -20 accuracies(\%) are reported. }
\label{table:PRID2011}
\begin{tabular}{| l| c | p{0.8cm}<{\centering} | p{2.2cm}<{\centering} |p{0.9cm}<{\centering} |p{1.3cm}<{\centering} p{1.3cm}<{\centering} p{1.3cm}<{\centering} p{1.3cm}<{\centering}| }
\hline
 \multirow{2}{*}{Methods} &  \multirow{2}{*}{Reference} & Deep  & Backbone & Optical    &   \multicolumn{4}{c|}{ PRID2011 } \\
\cline{6-9}
                          &   & model  & Network & Flow    &    top-1 & top-5 & top-10 & top-20    \\
\hline
1.~LFDA~\cite{C:LFDA}      & cvpr13& no  &--  & no & 43.7 & 72.8 & 81.7& 90.9 \\
2.~LADF~\cite{C:LADF}      & cvpr13& no  & -- & no & 47.3 & 75.5 & 82.7& 91.1 \\
3.~STFV3D~\cite{C:STFV3D}  & iccv15& no  &--  & no & 64.7 & 87.3 & 89.9& 92.0　\\
4.~TDL~\cite{C:TDL}        & cvpr16& no  &--  & no & 56.7 & 80.0 & 87.6& 93.6 \\
5.~CNN-RNN~\cite{C:CNN-RNN}& cvpr16& yes & SiameseNet~\cite{C:SiameseNet} & yes & 70.0 & 90.0 & 95.0& 97.0 \\
6.~CNN+XQDA~\cite{C:MARS}  & eccv16& yes & CaffeNet&no & 77.3 & 93.5 & --　& 99.3 \\
7.~TAM+SRM~\cite{C:JSTR}   & cvpr17& yes & CaffeNet&no & 79.4 & 94.4 & --　&　99.3 \\
8.~ASTPN~\cite{C:ASTPN}    & iccv17& yes & SiameseNet~\cite{C:ASTPN} & yes& 77.0 & 95.0 & 99.0& 99.0 \\
9.~QAN~\cite{C:QAN}        & cvpr17& yes & RPN~\cite{C:RPN} &no& 90.3 & 98.2 & 99.3& 100.0 \\
10.~RQEN~\cite{C:RQEN}     & aaai18& yes & GoogLeNet&no & 91.8 & \textcolor{blue}{\textbf{98.4}} & 99.3&　99.8　\\
11.~STAN~\cite{C:DiversReg}&cvpr18 & yes & ResNet50 &no & \textcolor{blue}{\textbf{93.2}} &  --  & --  & -- \\
12.~ST-Tubes~\cite{A:ST-Tubes}.&arxiv19&yes& ResNet50 &no& 78.0 & 89.0& 92.0&91.0 \\
\hline
12.~ours w/o optical       & -- & yes & ResNet50 & no & 92.0 & 98.0 &\textcolor{blue}{\textbf{100.0}} & \textcolor{blue}{\textbf{100.0}} \\
13.~ours w/ optical        & -- & yes & ResNet50 & yes & \textcolor{red}{\textbf{95.3}} & \textcolor{red}{\textbf{99.0}} & \textcolor{red}{\textbf{100.0}} & \textcolor{red}{\textbf{100.0}} \\
\hline
\end{tabular}
\end{center}
\end{table*}

\begin{table*}[t]
\begin{center}
\caption{ Performance comparison on the MARS by state-of-the-art methods. Our model is based on ResNet50. Top-1, -5, -20 accuracies(\%) and mAP(\%) are reported. }
\label{table:MARS}
\begin{tabular}{| l |c | p{0.8cm}<{\centering} | p{1.8cm}<{\centering} |p{0.9cm}<{\centering} |p{1.3cm}<{\centering} p{1.3cm}<{\centering} p{1.3cm}<{\centering} p{1.3cm}<{\centering}|}
\hline
 \multirow{2}{*}{Methods} & \multirow{2}{*}{Reference} & Deep    & Backbone & Optical & \multicolumn{4}{c|}{  MARS }  \\
\cline{6-9}
                          &                             & Model  & Network & Flow &  top-1 & top-5 & top-20 & mAP     \\
\hline
1.~CNN+Kiss.+MQ~\cite{C:MARS}        & eccv16 & yes & CaffeNet &no &68.3 & 82.6 &  89.4 & 49.3 \\
2.~Latent Parts~\cite{C:LatentPart}  & cvpr17 & yes & MSCAN~\cite{C:LatentPart}&no &71.8 & 86.6 &  93.0 & 56.1 \\
3.~TAM+SRM~\cite{C:JSTR}             & cvpr17 & yes & CaffeNet&no &70.6 & 90.0 &  \textcolor{blue}{97.6} & 50.7 \\
4.~QAN~\cite{C:QAN}                  & cvpr17 & yes & RPN~\cite{C:RPN} &no &73.7 & 84.9 &  91.6 & 51.7 \\
5.~K-reciprocal~\cite{C:k-reciprocal}& cvpr17 & yes & ResNet50 &no &73.9 & --   & 　--  & 68.5 \\
6.~TriNet~\cite{A:TriNet}            & arxiv17& yes & ResNet50 &no &79.8 & 91.4 &   --  & 67.7 \\
7.~RQEN~\cite{C:RQEN}                & aaai18 & yes & GoogLeNet&no &77.8 & 88.8 &  94.3 & 71.1 \\
8.~DuATM~\cite{C:DualAttention}      & cvpr18 & yes & DenseNet121  &no &78.7 & 90.9 &  95.8 & 62.3  \\
9.~STAN~\cite{C:DiversReg}           & cvpr18 & yes & ResNet50 &no &82.3 &  --  &  --   & 65.8  \\
10.~Part-Aligned~\cite{C:Part-Aligned}& eccv18& yes & InceptionV1&no& 84.7&94.4&97.5& 75.9 \\
11.~STA~\cite{A:STA}                  & arxiv19& yes & ResNet50 & no& 86.3& \textcolor{red}{\textbf{95.7}}& \textcolor{blue}{\textbf{98.1} }& \textcolor{red}{\textbf{80.8}} \\
\hline
12.~ours w/o optical                 &-- &yes & ResNet50& no&\textcolor{blue}{\textbf{86.6}} & 94.8&  97.1 &  76.7 \\
13.~ours w/ optical                  &-- &yes & ResNet50& yes &\textcolor{red}{\textbf{87.2}} & \textcolor{blue}{\textbf{95.2}} &  \textcolor{red}{\textbf{98.1}} & \textcolor{blue}{\textbf{77.2}} \\
\hline
\end{tabular}
\end{center}
\end{table*}

\subsection{Experimental Setting}\label{sec:expsetting}

\textbf{Datasets:} We evaluate the performance of proposed method on three well known video re-identification benchmarks: \textbf{the iLIDS-VID dataset}~\cite{C:iLIDS-VID}, \textbf{the PRID 2011 dataset}~\cite{C:PRID2011} and \textbf{the MARS dataset}~\cite{C:MARS}. (a) iLIDS-VID contains 600 image sequences of 300 pedestrians under two cameras.
Each image sequence has $23$ to $192$ frames. Both of the training and test set have $150$ identities.
(b) PRID is another standard benchmark for video re-identification. It consists of $300$ identities and each has $2$ image sequences. The length of sequences varies from $5$ to $675$.
(c) MARS is one of the largest video person re-identification dataset which contains $1,261$ different pedestrians and $20,715$ tracklets captured from $6$ cameras.
In this dataset, each person has one probe under each camera, resulting in $2,009$ probes in total.
The dataset is divided into training and test sets that contains $631$ and $630$ persons respectively.

\textbf{Evaluation Metric:} Two widely used evaluation metrics are employed for comparison. %
The first is the cumulative matching characteristic (CMC)~\cite{C:CMC}, which shows the probability of that a query identity appearing in different locations of the returned list~\cite{C:matket1501}.
In such case, re-id task is considered as a ranking problem and usually there is only one ground truth matching result for a given query.
Since the videos in MARS dataset are captured from $6$ camera, the ranking list may contain multiple matching results.
Thus, we also adopt mean average precision (mAP)~\cite{C:matket1501} to evaluate the performance in this dataset. In this case, the re-id problem is regarded as the retrieval problem.
For each query, we first calculate average precision (AP)~\cite{C:mAP} as follows,
\begin{equation}\label{eq:AP}
AP = \frac{1}{\sum_{i=1}^n r_i} \sum_{i=1}^n r_i \left(\frac{\sum_{j=1}^i}{i}\right),
\end{equation}
where $r_i$ is $1$ if the person in returned video $i$ has the same identity with the query, and $0$ otherwise.
$n$ is the total number of returned videos.
Then, the mean value of APs of all queries is calculated as the mAP, which considers both precision and recall of the method.

\begin{figure}
\includegraphics[width=0.5\linewidth]{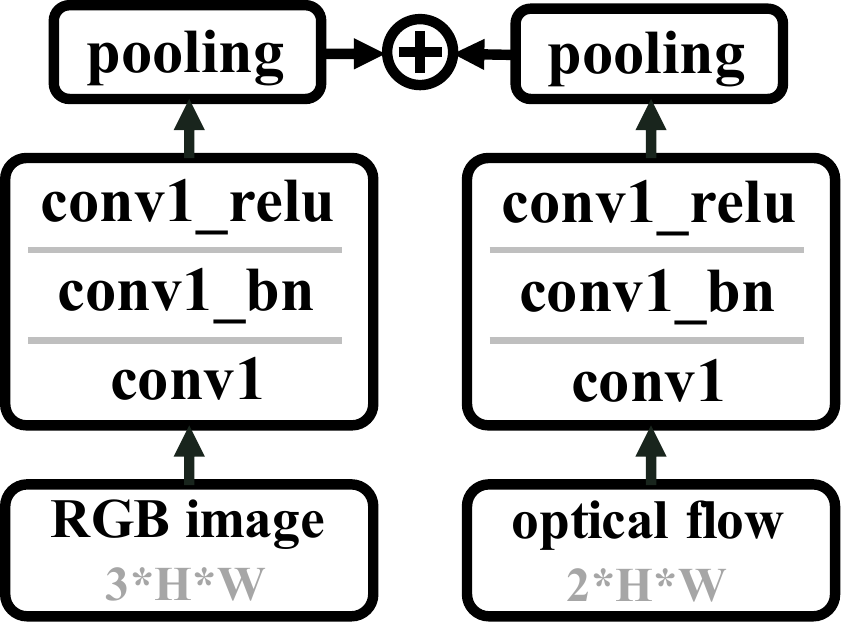}
\caption{Merging strategy of the RGB and optical flow branch. }
\label{fig:combine}
\end{figure}

\textbf{Optical Flow:} For further improving the performance of video re-id,
we use the optical flow~\cite{C:TSN,C:TwoStream} to extract the motion information from image sequence.
In practice, the dimension of input optical flow for each frame is $2*H*W$, where $2$ denotes the number of vertical and horizontal channels. $H$ and $W$ indicate the height and width of the map.
The value range of optical flow is scaled to $0$ to $255$.
Through one convolution layer (with BN and ReLU operation) and one pooling layer, the dimension of feature maps in optical branch becomes $64*\frac{1}{4}H*\frac{1}{4}W$, which is same as RGB branch.
Then an element-wise addition is applied to merge these two modalities, and the outputs are fed into the rest layers.
Fig.~\ref{fig:combine} illustrates the operation.

\subsection{Comparison with State-of-the-arts }\label{sec:SOTA}

%
We firstly report the comparison of proposed method with existing eleven state-of-the-art video person re-identification methods on iLIDS-VID dataset and PRID2011 dataset,
including LFDA~\cite{C:LFDA}, LADF~\cite{C:LADF}, STFV3D~\cite{C:STFV3D}, TDL~\cite{C:TDL}, CNN-RNN~\cite{C:CNN-RNN}, CNN+XQDA~\cite{C:MARS}, TAM+SRM~\cite{C:JSTR}, ASTPN~\cite{C:ASTPN}, QAN~\cite{C:QAN}, RQEN~\cite{C:RQEN}, STAN~\cite{C:DiversReg} ST-Tubes~\cite{A:ST-Tubes}.
The first four methods are traditional methods without using deep models, while the others adopt deep neural networks to extract the feature representation of each frame.
We use ResNet50~\cite{C:ResNet} as the basic model of proposed SCAN.
Following~\cite{C:ASTPN}, each dataset is randomly split into $50\%$ of identities for training and others for testing.
All experiments are repeated $10$ times with different train/test splits, and the averaged results are reported~\cite{C:ASTPN}.
As shown in Table~\ref{table:iLIDS-VID} and Table~\ref{table:PRID2011},
our method achieves state of the art $88.0\%$ and $95.3\%$ top-1 accuracy on iLIDS-VID and PRID2011,
outperforming the existing best method STAN~\cite{C:DiversReg} with $7.8\%$ and $2.1\%$, respectively.
%
%

For the iLIDS-VID and PRID2011 dataset, only CNN-RNN~\cite{C:CNN-RNN} and ASTPN~\cite{C:ASTPN} adopt optical flow to capture the motion information.
According to Table~\ref{table:iLIDS-VID}, Table~\ref{table:PRID2011} and Fig.\ref{fig:differnetDepth},
when we use AlexNet as the backbone network whose depth is close to their SiameseNet ($3\sim5$ convolution layers), our method (\textit{i.e.} with optical flow), which achieves top-1 accuracy $69.8\%$ and $85\%$ on iLIDS-VID and PRID2011 dataset, still outperforms the above two approaches.

To further demonstrate the effectiveness of SCAN on the data captured from multiple camera views, we compare it with state-of-the-arts on MARS dataset, including CNN+Kissme+MQ~\cite{C:MARS}, Latent Parts~\cite{C:LatentPart}, TAM+SRM~\cite{C:JSTR}, QAN~\cite{C:QAN}, K-Recip.~\cite{C:k-reciprocal}, TriNet~\cite{A:TriNet}, RQEN~\cite{C:RQEN}, DuATM~\cite{C:DualAttention}, STAN~\cite{C:DiversReg}, Part-Aligned~\cite{C:Part-Aligned} and STA~\cite{A:STA}.
Table~\ref{table:MARS} reports the retrieval results.
Our method achieves $87.2\%$ and $86.6\%$ top-1 accuracy with and without using optical flow, which surpasses all existing work.
For the mAP, top-5 and top-20 accuracy, our method achieves competitive results compared to the most recent
work in Table~\ref{table:MARS},
implying that the proposed SCAN is also suitable for large-scale video re-identification task.
It is noteworthy that the STA~\cite{A:STA} is the only method that goes beyond the performance of ours.
However, it need to calculate the spatial and temporal attention scores simultaneously, which brings in more
computational cost in the training and test phase.
%

The reason that the proposed method outperforming most of the state-of-the-arts can be summarized into three aspects.
Firstly, compared with recent proposed temporal modeling methods~\cite{C:CNN-RNN,C:ASTPN}, which usually adopt RNN or LSTM to generate video representations, the self-attention mechanism reweights the frames in each video and refines the intra-video representations according to the updated weights.
Such scheme reduces the impact of noise frames with variant appearance on intra-person feature representations.
Secondly, the collaborative-attention mechanism effectively captures the discriminative frames to learn inter-video representations, achieving more accurate similarity between probe and gallery image sequences.
In addition, transforming such matching problem into a binary classification problem makes the optimization of such problem easier than previous pair-wise loss or triplet loss based schemes, which need to predefine a suitable margin threshold to leverage convergence speed and over-fitting problem.
At last, a dense clip segmentation strategy produces many hard positive and hard negative pairs to learn the similarity between two videos, making the proposed model more robust in the test phase.

\begin{table*}[t]
\begin{center}
\caption{ Comparison of different temporal modeling methods. Top-1, -5 accuracies(\%) and mAP(\%) are reported. }
\label{table:Ablation}
\begin{tabular}{| l || p{1cm}<{\centering}  p{1cm}<{\centering}  p{1cm}<{\centering} | p{1cm}<{\centering}   p{1cm}<{\centering}  p{1cm}<{\centering} | p{1cm}<{\centering}   p{1cm}<{\centering}  p{1cm}<{\centering} |}
\hline
 \multirow{2}{*}{Seetings} &    \multicolumn{3}{c|}{ iLIDS-VID } & \multicolumn{3}{c|}{  PRID2011 } & \multicolumn{3}{c|}{  MARS } \\
\cline{2-10}
                          &    top-1 & top-5 & mAP & 　top-1 & top-5　& mAP    & 　top-1 & top-5　& mAP  \\
\hline
1. ave. pooling               & 81.3 &  95.3 & 84.5 & 92.0 &\textbf{100.0} &93.6  &　83.4 & 93.2 & 75.1 \\
2. max pooling               & 82.0 &  96.0 & 85.2 & 92.0 &100.0 &93.6   &　83.6 & 93.1 & 75.0 \\
3. SAN only                  & 82.7 &  97.3 & 85.7 & 92.0 &99.0　&93.8   &　84.2 & 94.7 & 75.4 \\
4. CAN only                  & 83.3 &  96.0 & 84.3 & 92.0 & 100.0&93.5   &　85.4 & 95.1 & 75.7 \\
5. SCAN \textit{single path}  & 86.7  & 96.7 &  88.5 & 93.4 & 99.0 & 94.6  & 86.1　& 95.1 & 76.3 \\
6. SCAN \textit{same fc layer}  & 86.6 & 96.0 & 90.6 & 93.4 &  98.6 & 95.6 & 85.8 & 94.9 & 75.6 \\
7. SCAN \textit{dot product}    & 86.0  & \textbf{97.0} & \textbf{90.7} & 93.6 & 98.2 &95.7  & 86.5 & 94.8 &76.7\\
8. our full model            & \textbf{88.0} &  96.7& 89.9 & \textbf{95.3} & 99.0 &\textbf{95.8} &　\textbf{87.2}& \textbf{95.2 }& \textbf{77.2 }\\
\hline
\end{tabular}
\end{center}
\end{table*}


\begin{table*}[t]
\begin{center}
\caption{ Setting the length of video clips is critical. The `stride' denotes the overlap of two consecutive clips. Top-1, -5 accuracies(\%) and mAP(\%) are reported. }
\label{table:cliplength}
\begin{tabular}{| l || p{1cm}<{\centering}  p{1cm}<{\centering}  p{1cm}<{\centering} | p{1cm}<{\centering}  p{1cm}<{\centering}  p{1cm}<{\centering} | p{1cm}<{\centering}  p{1cm}<{\centering}  p{1cm}<{\centering} |}
\hline
 \multirow{2}{*}{Seetings} &    \multicolumn{3}{c|}{ iLIDS-VID } & \multicolumn{3}{c|}{  PRID2011 } & \multicolumn{3}{c|}{  MARS }\\
\cline{2-10}
                          & top-1 &  top-5 & mAP      &　top-1　& top-5 & mAP  &　top-1　& top-5 & mAP  \\
\hline
a. 10-frames 3-stride      &   87.6  &  96.6 &  89.9 &  93.4 & 98.4 & 95.7 & 86.0& 94.3& 76.1\\
b. 10-frames 5-stride              & \textbf{88.0 }& \textbf{96.7} &\textbf{89.9}       & \textbf{95.3} & 99.0 &\textbf{95.8} & \textbf{87.2}& \textbf{95.2 }& \textbf{77.2 }  \\
c. 16-frames 8-stride              & 85.3 & 95.3 &87.8       & 94.0 & 99.0  &94.5  & 86.3 & 95.1 & 75.8 \\
d. 20-frames 10-stride             & 78.0 & 96.0 &81.4       & 89.0 & \textbf{100.0} &91.3  & 85.5 & 94.8 & 75.4 \\
\hline
\end{tabular}
\end{center}
\end{table*}

\subsection{Ablation Study }\label{sec:ablation}

%
%
%

To investigate the efficacy of proposed SCAN, we conduct ablation experiments on iLIDS-VID, PRID2011 and MARS dataset. The average pooling over temporal dimension is used to be our baseline model and ResNet50~\cite{C:ResNet} is adopted as the bottom Convolutional Neural Networks if not specified.
The overall results are shown on Table~\ref{table:Ablation}.
We also consider the impact of the cutting length of video clips. The comparison results are shown in Table~\ref{table:cliplength}.


\textbf{Instantiations}. We compare our full model with seven simplified settings, including
\begin{itemize}
\item using the average pooling over temporal dimension to calculate the feature representation of both the probe and gallery sequences;
\item using max pooling to replace average pooling in above;
\item using Self Attention Network (SAN) to compute probe and gallery video features separately;
\item using average pooling to obtain the video-level feature representation firstly, and  using Collaborative Attention Network (CAN) to reconstruct probe and gallery video representations;
\item using SAN to calculate probe video feature, followed by employing such feature representation to reconstruct gallery video representation by CAN, and setting can be viewed as a \textit{single-path} variant of the proposed SCAN;
\item using the same FC layer in SAN and CAN, \textit{i.e.} the $fc_1$ and $fc_2$ share the parameters, and such setting is named as  \textit{same fc layer} in the rest article.

\item using dot product to instead Hadamard product to calculate the correlation weight in Eqn.(\ref{eq:scr}) and Eqn.(\ref{eq:ccr}), and this setting is denoted as \textit{dot product} for comparison.
\end{itemize}
%

According to Table~\ref{table:Ablation}, we have several important findings.
First, the baseline model (\textit{i.e.} ave. pooling) has already outperformed several state-of-the-art methods with a margin.
It demonstrates the effectiveness of proposed pipeline, including clip segmentation and binary cross-entropy loss.
Second, the matching accuracy achieves a slightly improvement when only using SAN or CAN for temporal modeling, but \textit{single path} SCAN outperforms the baseline with a margin.
It suggests that the SAN and CAN modules are coupled when aligning the discriminative frames in the probe and gallery image sequences.
Third, the performance of the \textit{single-path} SCAN is less than our full model, reflecting the importance of generalized similarity representation between probe and gallery sequences in the matching problem.
At last, when sharing the parameters in the $fc_1$ and $fc_2$ layer or replacing Hadamard product with dot product in Eqn.(\ref{eq:scr}) and Eqn.(\ref{eq:ccr}), the accuracy of SCAN decreases on all of three datasets, but still outperforms the most of state-of-the-arts.
It demonstrates the robustness of proposed framework. Meanwhile, we suggest using different FC layers and Hadamard product to further promote the performance of our method in practice.

\textbf{Video clip with different length}. Using clip segmentation strategy can well improve the performance of video recognition~\cite{C:S3D}.
In this paper, we also investigate the performance of the SCAN model using different length of video clips. We cut the input image sequence into several clips with $10$ frames, $16$ frames and $20$ frames,
and the number of overlapped frames (\textit{i.e.}, the stride of the sliding window over the temporal dimension) is set as $3$, $5$, $8$ and $10$, respectively .
In Table~\ref{table:cliplength}, the setting with ($10$ frames, $5$ stride) achieves the best performance over all of the evaluation metrics.
We can also observe that as the clip length grows, the accuracy drops gradually.
It demonstrates the cutting strategy can provide more diverse pairs in the minibatch, which increases the model capacity effectively.

\begin{figure}[t]
\centering
\includegraphics[width=\linewidth]{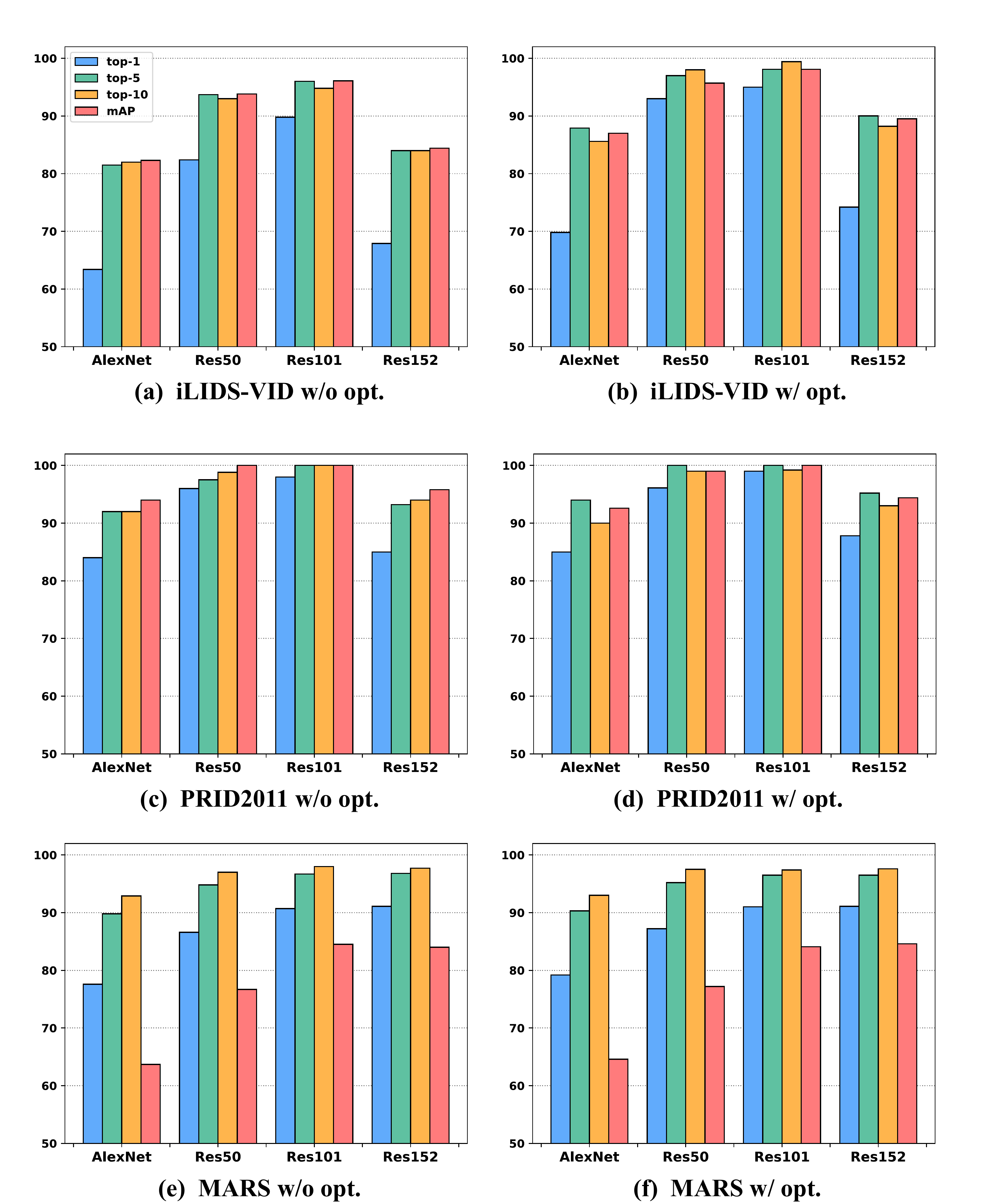}
\caption{The performance of the models using different depth on iLIDS-VID, PRID2011 and MARS dataset. w/o and w/ indicate with and without the optical flow, respectively.  Top-1, -5 and -10 accuracies(\%) and mAP(\%) are reported.
}
\label{fig:differnetDepth}
\end{figure}

\textbf{The depth of neural networks}. Deeper neural networks have been beneficial to image classification task. To further analyze proposed SCAN, we also conduct experiments for different depths of pre-trained CNN.
We test AlexNet~\cite{C:AlexNet} and ResNet~\cite{C:ResNet} with three different depths, \textit{i.e.}, $50$, $101$ and $152$ layers.
As shown in Fig.~\ref{fig:differnetDepth}, ResNet has an obvious advantage in re-id task.
With the same setting, increasing the depth of ResNet can only achieve slight improvement when optical flow is ignored.
It means when the depth is larger than $50$, our method is not very sensitive to the depth of the network.
On the other hand, when we take the optical flow into consideration, the accuracies of deeper ResNets have declined a little.
This may be caused by the suboptimal optical flow merging strategy in the bottom layer of the network.

According to Fig.~\ref{fig:differnetDepth}, we also find that the improvement by using optical flow on MARS is little, e.g. the top-1 accuracy by using ResNet50 is $87.2\%$ vs. $86.6\%$ with and without optical flow, and the mAP is $77.2\%$ vs. $76.7\%$.
Such experiment result is consistent with the opinion in~\cite{C:MARS} that motion features (e.g. optical flow, HOG3D, GEI and so on) have poor performance on MARS.
The reason can be summarized in two-fold.
Firstly, as one of the largest datasets for video-based re-id, MARS contains many pedestrians sharing similar motion feature, thus it is more difficult than the other two datasets to distinguish different persons based on motions.
Secondly, since the samples are captured by six cameras in MARS, motion of the same identity may undergo significant variations due to pedestrian pose change, so the motion-based feature can provide less information to discriminate the same person with various motion views.

\begin{figure}[t]
\centering
\includegraphics[width=\linewidth]{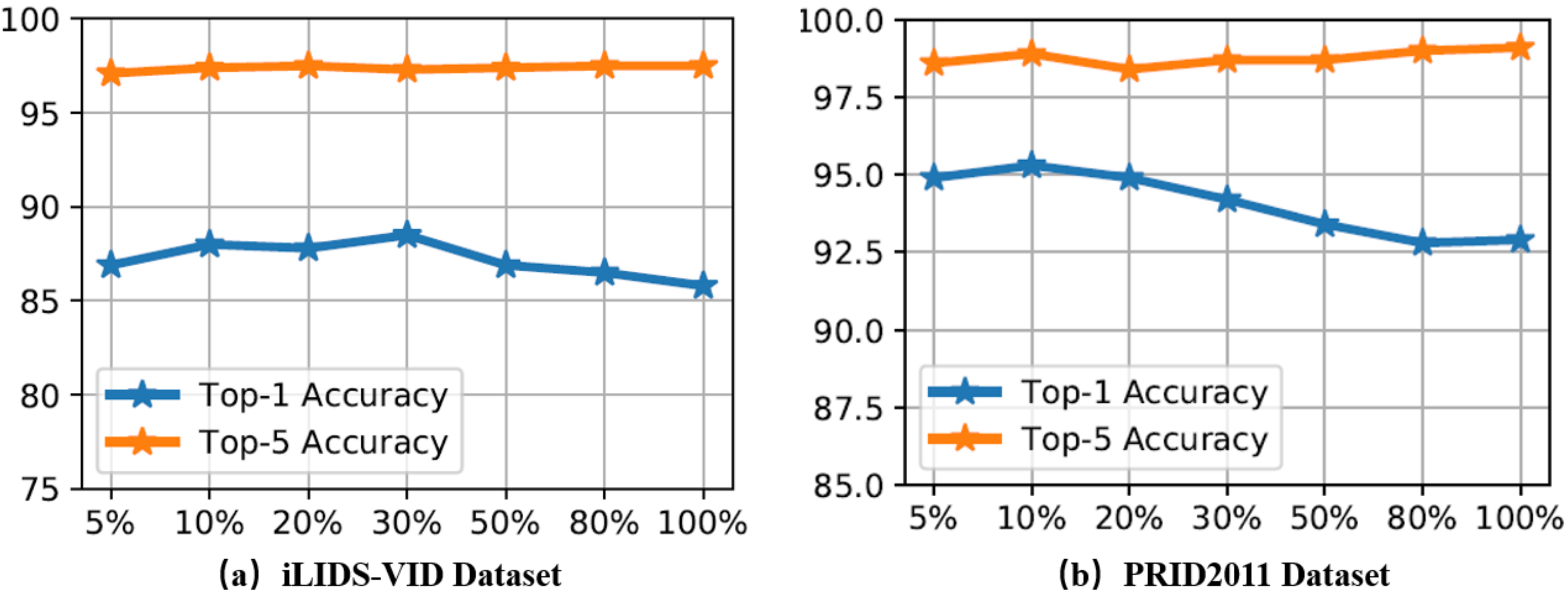}
\caption{ Top-1 and top-5 accuracies(\%) on iLIDS-VID and PRID2011 dataset by using varying ensemble rates.}
\label{fig:SelectRate}
\end{figure}

\textbf{Different ensemble rates}. In the test phase, the proposed model averages the $k\%$ probe-gallery clip-pairs with the highest matching scores to estimate the final matching results. $k\%$ indicates the ensemble rate.
Such ensemble strategy can effectively inhibit the effects of clip-pairs with very low matching confidences.
To investigate the effect of ensemble rate on matching accuracies, we adopt different ensemble rates to evaluate the performance.
Fig.~\ref{fig:SelectRate} reports the top-1 and top-5 accuracies on iLIDS-VID and PRID2011 dataset by using $7$ ensemble rates.
It shows that different datasets have their own appropriate ensemble rate.
According to a comprehensive analysis of the results on these two datasets, we set the ensemble rate as $10\%$ for all of the experiments in this paper.

\begin{table*}[t]
\begin{center}
\caption{ Cross dataset matching results. The first row indicates the name of test dataset. The fist and second column indicate the method and the training set, respectively. Top-1, -5 -10 and -20 accuracies(\%) are reported.}
\label{table:crossdataset}
\begin{tabular}{ l | c | cccc | cccc   }
\hline
 \multirow{2}{*}{Method} & \multirow{2}{*}{Training Set} & \multicolumn{4}{c|}{ iLIDS-VID } & \multicolumn{4}{c}{  PRID2011 } \\
\cline{3-10}
                       &   &top-1 & top-5 & top-10 & top-20   & top1 　& top-5　& top-10 & top-20 \\
\hline
CNN-RNN~\cite{C:CNN-RNN}               &  \multirow{2}{*}{iLIDS-VID}  & - & - & - & -        &28.0  & 57.0 & 69.0& 81.0   \\
ASTPN~\cite{C:ASTPN}                   &   & - & - & - & -        & 30.0 & 58.0 & 71.0& 85.0   \\
\hline
                   & iLIDS-VID  & - & - & - & -        & 29.5 & 59.4 & - & 82.2   \\
TRL~\cite{A:TRL}   & PRID2011  & 8.9 & 22.8 & - & 48.8        & - & - & -& -   \\
                   & MARS  & 18.1 & 30.8 & - & 59.3        & 35.2 & 69.6 & -& 89.3   \\
\hline
                    & iLIDS-VID  & - & - & - & -        & 42.8 & 71.6 & 80.2& 88.9   \\
our full model      & PRID2011  & 9.7 & 27.5 & 36.9 &  48.6       & - & - & -& -    \\
                    & MARS      & 19.3 & 46.7 & 56.2 &  66.0       & 46.0 & 69.0 & 82.0&  92.0  \\
\hline
\end{tabular}
\end{center}
\end{table*}

\subsection{Cross-dataset Generalization}\label{sec:crossdataset}

Due to the variety conditions in the process of data collection, the data distributions of different datasets may have great bias.
The performance of the model trained on one dataset may drop a lot on another one.
To evaluate the generalization ability of proposed model, as well as to understand the difference of standard benchmarks, we conduct cross-dataset validation with two settings.
Following~\cite{A:TRL}, for the first setting, the re-id model is trained on the large-scale MARS dataset and tested on the iLIDS-VID and PRID2011.
The second one is training on iLIDS-VID or PRID2011, and testing on the other dataset.
Table~\ref{table:crossdataset} shows the top-1,-5,-10,-20 accuracies.

According to Table~\ref{table:crossdataset}, models trained on MARS dataset achieve better generalization performance on both iLIDS-VID and PRID2011.
It shows the  benefits of large-scale datasets in training the models with better generalization ability.
However, the accuracies of all methods still decline sharply, which demonstrates the disparities of the data distributions between MARS and the other two datasets.
When trained on MARS dataset, our model achieves $19.3\%$ and $46\%$ of top-1 accuracies on iLIDS-VID and PRID2011, exceeding all compared methods and outperforming~\cite{A:TRL} with $1.2\%$ and $9.8\%$, which proves certain generality of proposed SCAN by using large-scale dataset.

For the second setting, when the model is trained on one of the two small datasets, the matching accuracies drop a lot on another dataset compared with Table~\ref{table:iLIDS-VID} and~\ref{table:PRID2011}.
Specially, when training on PRID2011 dataset, our model achieves best cross-dataset performance on iLIDS-VID with $9.7\%$ top-1 accuracy.
If we adopt iLIDS-VID as the training set, the top-1 accuracy of SCAN on PRID2011 is $42.8\%$, outperforming the other baselines.
But it declines more than half of the accuracy compared with our best model in Table~\ref{table:PRID2011}.
The above results suggest that the proposed SCAN has certain generalization for the small dataset cross testing compared with other baselines, but still need to be further improved.
They also demonstrate that iLIDS-VID dataset is more diverse and challenging than
PRID2011 dataset.

\section{Conclusions}
\label{sec:conclusion}

In this paper, we propose a temporal oriented similarity measurement to further promote the performance of video-based person re-identification.
A novel module named Self-and-Collaborative Attention Network (SCAN),
which integrates frame selection, temporal pooling and similarity measurement into a simple but effective module, is introduced to pursuit this goal.
Different from previous deep metric learning methods that project the video-level representations into a common feature space for similarity measurement,
SCAN is a well designed non-parametric module which can align the discriminative frames between probe and gallery videos in the `\textit{frame space}'.
Such a scheme is significant in the temporal-based matching problem, as well as other video-based vision problem.
Extensive experiments demonstrate that the proposed SCAN outperforms the state-of-the-arts on top-1 accuracy.

Several directions can be considered to further improve our model.
First, extending SCAN into the spatial-temporal dimension is an intuitive idea.
Second, how to efficiently integrate the multi-modality information, e.g., RGB and optical flow, into a single framework is still an open issue.
Third, combining proposed method with other visual tasks, such as video object detection or video-based instance segmentation, is also an exciting research direction.
At last, combining our model with unsupervised methods is also a potential direction.

\bibliographystyle{IEEEtran}
\bibliography{IEEEtran}


%




\ifCLASSOPTIONcaptionsoff
  \newpage
\fi

\end{document}